\documentclass{article}
\pdfoutput=1 %PER ARXIV
\usepackage{amssymb}
\usepackage{amsmath}
\usepackage{lipsum}
\usepackage[dvipsnames]{xcolor}
\usepackage{cleveref}
\usepackage{multirow}
\usepackage{graphicx}
\usepackage{subcaption}
\usepackage{url}
\usepackage{algorithmicx}
\usepackage{algpseudocode}
\usepackage[linesnumbered,ruled,vlined]{algorithm2e}
\usepackage{cleveref}
\renewcommand{\vec}[1]{\mathbf{#1}}

\let\citep\cite

\title{IMPACTX: improving model performance by appropriately constraining the training with teacher explanations}
\author{Andrea Apicella, Salvatore Giugliano, Francesco Isgrò, Andrea Pollastro, Roberto Prevete\\
\textit{\small Department of Electrical Engineering and Information Technology, University of Naples Federico II}}
\date{}

\begin{document}
\maketitle
\begin{abstract}
%Modern Artificial Intelligence (AI) systems, especially Deep Learning (DL) models, poses challenges in understanding their inner workings by AI researchers. eXplainable Artificial Intelligence (XAI) inspects internal mechanisms of AI models providing explanations about their decisions. While current XAI research predominantly concentrates on explaining AI systems, there is a growing interest in using XAI techniques to automatically improve the performance of AI systems themselves.
\let\thefootnote\relax\footnotetext{This paper has been published on Artificial Intelligence Review journal. Please refer to the published version at \url{https://link.springer.com/article/10.1007/s10462-026-11564-z}}The eXplainable Artificial Intelligence (XAI) research predominantly concentrates to provide explainations about AI model decisions, especially Deep Learning (DL) models.  However, there is a growing interest in using XAI techniques to automatically improve the performance of the AI systems themselves.

This paper proposes \textit{IMPACTX}, a novel approach that leverages XAI as a fully automated attention mechanism, without requiring external knowledge or human feedback. Experimental results show that IMPACTX has improved performance respect to the standalone ML model by integrating an attention mechanism based an XAI method outputs during the model training. Furthermore, IMPACTX directly provides proper feature attribution maps for the model's decisions, without relying on external XAI methods during the inference process.

Our proposal is evaluated using three widely recognized DL models (EfficientNet-B2, MobileNet, and LeNet-5) along with three standard image datasets: CIFAR-10, CIFAR-100, and STL-10. The results show that IMPACTX consistently improves the performance of all the inspected DL models across all evaluated datasets, and it directly provides appropriate explanations for its responses.
\end{abstract}

%\begin{keyword}

%keyword one \sep keyword two
%\end{keyword}

%\end{frontmatter}


\section{Introduction}
\label{sec:intro}
The inner workings of modern Machine Learning (ML) and Deep Learning (DL) approaches are often opaque, leaving AI scientists to interpret the reason behind the provided outputs. eXplainable Artificial Intelligence (XAI) aims to provide insights into the internal mechanisms of AI models and/or the motivations behind their decisions, helping users to understand the outcomes produced by ML models. XAI approaches are adopted in several ML tasks applied to various types of inputs, including images \cite{ribeiro2016should,apicella2019explaining,montavon2019layer}, natural language processing, clinical support systems \cite{schoonderwoerd2021human,annuzzi2023impact}, and more.
However, it's worth noting that while a considerable portion of the existing XAI literature focuses on providing explanations for AI systems \cite{lipton2018mythos,miller2019explanation,arrieta2019explainable}, another emerging aim of the XAI studies is to improve the ML systems themselves \cite{weber2022beyond}. Current literature on this topic often presents approaches that require human interactions, such as \cite{teso2019explanatory,schramowski2020making,hagos2022impact,ijcai2017p371,mitsuhara2019embedding,zunino2021explainable,selvaraju2019taking}. %for instance, based on the explanations provided, humans may choose to eliminate specific data from the dataset (e.g., Explanatory Interactive Learning, \cite{schramowski2020making}). 
In these works, the human role can be viewed as a method of manually directing attention, guiding the model towards particular input features.

More recently, however, there has been a growing emphasis on using XAI techniques to autonomously improve machine learning systems as part of the training stage, without requiring a direct human intervention \cite{weber2022beyond,sun2022utilizing,liu2023icel,sun2021explanation}.
The basic assumption is that explanations of the model outputs can improve the training of the ML systems to find better solutions. In particular, the employed XAI methods can be perceived as a type of attention mechanism \cite{vaswani2017attention} to able to lead the model towards improved performance (see for example \cite{fukui2019attention}). 
This kind of approaches typically involved altering the training procedure or the model's loss function to emphasize relevant information \cite{yeom2021pruning,ijcai2017p371,liu-avci-2019-incorporating}. %As a result, a significant portion of pre-trained ML models available on large and common datasets cannot be exploited or has to be fine-tuned using these proposed learning strategies. In state-of-the-art DNNs, this process can be computationally demanding due to the substantial number of model parameters involved.

In this paper, we introduce IMPACTX (Improving Model Performance by Appropriately Predicting CorrecT eXplanations), a novel approach that leverages AI as an attention mechanism in a fully automated manner, without requiring external knowledge or human feedback. IMPACTX achieves two primary goals: (i) enhancing the performance of an ML model by integrating an attention mechanism trained on the output of an XAI method, 
%during the model training, 
and  (ii) directly providing appropriate explanations for the model's decisions, making the whole framework inherently \textit{self-explanatory}. \textcolor{black}{
It is important to note that this self-explanatory capability is acquired at the end of the learning phase, whereas during the learning phase, IMPACTX relies on an external XAI method to refine and enhance the model's self-explanatory ability.}
In a nutshell, this is achieved through a dual-branch model: the first branch consists of a feature extractor and a classifier, and the second branch is an Encoder-Decoder scheme that encodes useful information about relevant input features for classification. The interaction between these two branches guides the classifier during the classification task while simultaneously providing explanations for the given output by the Encoder-Decoder branch.
As we will discuss in section \ref{sec:related}, IMPACTX is different from other approaches not requiring direct human intervention because IMPACTX uses both augmented intermediate features and augmented loss function, and it is a model-agnostic approach.
%In summary, IMPACTX has the following properties: 1) It integrates machine learning tasks with XAI, showing that XAI is not only effective in explaining ML to humans but also in self-improving ML without human intervention. 2) It delivers valid explanations for its outputs at inference time, without relying on external XAI methods after the inference process.

To evaluate our proposal, we experimentally assess the ability of IMPACTX to both improve the performance of ML models and provide meaningful explanations in the form of attribution maps. This experimental evaluation is conducted on three widely recognized DL models: EfficientNet-B2, MobileNet, and LeNet-5. All the experiments are conducted using  three standard image datasets: CIFAR-10, CIFAR-100, and STL-10. %MobileNet and EfficientNet-B2 are initialised with the weights of ImageNet and Noisy Student, respectively. %explanations are generated using SHAP (SHapley Additive exPlanations) \cite{NIPS2017_7062}, a widely recognized XAI method.

The paper is organised as follows: \cref{sec:related} provides an overview of related works that use XAI to enhance machine learning systems. Our proposed method is detailed in \cref{sec:method}, followed by an outline of the experimental assessment in \cref{sec:exp}. \Cref{sec:res} presents the results of the experimental assessment, while \cref{sec:discussion} comprehensively discusses these results, and concluding with final remarks in \cref{sec:conclusion}.

\section{Related works}
\label{sec:related}
Different kinds of explanations have been proposed to address the explainability problem \citep{bach2015,dosovitskiy2016,ribeiro2016should,NIPS2017_7062,selvaraju2017grad}, depending on both the AI system being explained and the XAI method adopted. A very common approach consists of providing visual-based explanations in terms of input feature importance scores, known as \textit{attribution } or \textit{relevance } maps. Examples of this approach include Activation Maximization (AM) \citep{erhan2009}, Layer-Wise Relevance Propagation (LRP) \citep{bach2015}, Deep Taylor Decomposition \citep{binder2016,montavon2017_2}, Deconvolutional Network \citep{zeiler2011}, Up-convolutional Network \citep{zeiler2014,dosovitskiy2016}, and SHAP method\citep{NIPS2017_7062}. 

Importantly, in the context of enhancing ML systems
by XAI methods, these methods can be viewed as a way of exploiting external knowledge which can be provided by human intervention, such as the annotation of  attribution maps, or in a fully automated manner without the involvement of a human operator (see, for example, \cite{hu-etal-2016-harnessing,apicella2018integration}).
However, as already mentioned in section \ref{sec:intro}, we note that a significant area of the literature focuses on improving ML models through human intervention in the context of XAI, as evidenced in several studies \citep{teso2019explanatory, schramowski2020making, hagos2022impact, ijcai2017p371, mitsuhara2019embedding, zunino2021explainable, selvaraju2019taking}.
%in terms of explanations from data to enhance Machine Learning (ML) systems, as shown in studies such as \citep{hu-etal-2016-harnessing,apicella2018integration}. This external knowledge can be provided by human intervention, such as the annotation of  attribution maps, or in a fully automated manner without the involvement of a human operator. 
This aspect can also be deduced by \cite{weber2022beyond}, a recent survey of research works using XAI methods to improve ML systems. Moreover, \cite{weber2022beyond} discusses efforts to improve ML models along different dimensions such as \textit{performance}, \textit{convergence}, \textit{robustness}, and \textit{efficiency}. For improving ML model performance with XAI, \cite{weber2022beyond} isolates four main approaches: % in realtà sono 5, ma nella categoria 5 non ci sono metodi che migliorano le performance
i) \textit{augmenting the data}, explanations are utilized to generate artificial samples or to alter the distribution of data that provide insights into undesirable behaviors \citep{teso2019explanatory, schramowski2020making}; 
ii) \textit{augmenting the intermediate features}, by measuring feature importance through explanations, this information can be effectively used to scale, mask, or transform intermediate features to improve the model \citep{anders2022finding, fukui2019attention, apicella2023strategies, apicella2023shap};
iii) \textit{augmenting the loss}, additional regularization terms based on explanations are incorporated into the loss training function \citep{ijcai2017p371, ismail2021improving, liu2023icel}; 
iv) \textit{augmenting the gradient}, information about the importance of the features provided by explanations can also be applied during the backpropagation pass by changing the gradient update \citep{ha2019improvement}.

As described in section \ref{sec:method}, the proposed method, IMPACTX, focuses on improving the performance of an ML system and it is related to category ii) ‘augmenting the intermediate features’ and iii) 'augmenting the loss' in a fully automated manner without involvement of human knowledge in the learning step. 

In this context, i.e., to improve ML system performance by XAI methods in an automated manner without involvement of human knowledge in the learning step, \cite{bargal2021guided} propose Guided Zoom to improve the performance of models on visual data, especially on fine-grained classification tasks (i.e., where the differences between classes are subtle),  This method aims to improve fine-grained classification decisions by comparing consistency of the evidence for the incoming inputs with the evidence seen for correct classifications during the training. %refine the model's predictions by comparing them with the tests used at the time of training that led to correct decisions. In particular, the contrastive Excitation Backprop \citep{zhang2018top} is used as a grounding method.
\cite{mahapatra2021interpretability} (IDEAL) use local explanations in medical image analysis to select the most informative samples. %They compute an informativeness score for each sample based on its explanation and rank the samples in descending order. New samples for labelling and training are then selected from this ranking. Their method, IDEAL, shows superior performance over other sample selection criteria, requiring less data and fewer iterations to achieve comparable or better results.
The Guided Zoom \citep{bargal2021guided} and IDEAL \citep{mahapatra2021interpretability} methods differ from our approach in that they use XAI to augment the data. 
%%In contrast, as we will discuss in sec. \ref{sec:method}, IMPACTX employs XAI to augment intermediate features and loss function.
% gradient aug.
In \cite{ha2019improvement} variants of parameter gradient augmentation based on LRP \citep{montavon2019layer} are introduced.% able to the direction of each learning step and affecting the performance. %This method augments the gradient solely during the backward step, whereas our approach also performs a type of augmentation during the forward step.
% metodi "masks"
Moreover, a number of methods \citep{schiller2019relevance, zunino2021excitation,ismail2021improving, apicella2023strategies} mask the input features using XAI techniques with the aim of improving performance. %%In a different way, our method does not involve masking adopting XAI, but XAI methods are exploited to augment intermediate features and loss function. %exploits the latent coding of explanations using XAI (see Sec. \ref{sec:method}).

In \cite{sun2021explanation} LRP explanations \citep{montavon2019layer} led an ML model to focus on the important features during the training stage of a few-shot classification task. %
%The Right for the Right Reasons (RRR) \citep{ijcai2017p371} method mentioned above also has an approach when human annotations are not available. In this case, they use rules to adapt the explanations to an ensemble of models. 
% 
\cite{liu2023icel} incorporate a contrastive comparison between heatmaps of two different categories as part of the loss function. % starting from the intuitive assumption that human beings distinguish different object through differential details, the authors of . %
In \cite{sun2022utilizing}, a retraining strategy is proposed that uses Shapley values \citep{NIPS2017_7062} to assign specific training weights to misclassified data samples, thereby improving model predictions. %The method has been tested on a single known model and dataset, whereas our approach has been tested on multiple models and datasets. 

We underline that all the previously described approach differ from our proposal insofar as we use both augmented intermediate features and augmented loss function.

By contrast, the works described in \cite{fukui2019attention,apicella2023shap} appear to share a number of common aspects with our proposal.
% più simili, abn, wae.. e??
In particular, \cite{fukui2019attention} uses, similarly to what we propose, an additional attention branch, the Attention Branch Network (ABN), to improve deep Convolutional Neural Networks (CNN). The ABN extends Class Activation Mapping (CAM) \citep{zhou2016learning} to generate explanations and use this knowledge to improve performance. However, ABN differs from our proposal since it is a not model-agnostic approach (i.e. it requires access to the internal mechanisms of the ML model and is dedicated to CNN models), while our approach can be applied to any type of ML models without knowledge about the internal mechanisms of the model itself. Moreover, ABN employs an input mask-based strategy with attention masks generated during the training phase, while our approach augments intermediate features and the loss training function exploiting XAI explanations. %and IMPACTX trains on explanations obtained in an earlier phase using the XAI method SHAP \citep{NIPS2017_7062}. 
Similarly, in \cite{apicella2023shap} we proposed an architecture composed of an additional attention branch in terms of an explanation encoder designed to augment the intermediate features through Weighted Average Explanations (WAE) obtained by Shapley values \citep{NIPS2017_7062}.

Summarizing, since the literature works which seem more similar to our proposal are ABN \citep{fukui2019attention} and WAE \citep{apicella2023shap}, a direct comparison between ABN, WAE and the proposal of this work will be made in sec. \ref{sec:res}.

%
 %\roberto{Ma noi non modifichiamo anche la loss dicendo che il ramo LEP-D deve produrre una buona spiegazione??}. \sal{ragionandoci meglio, per come la intende Weber in cui si fa loss augumentation quando si sfrutta la spiegazione nella loss, sì. Avevo considerato ABN, che per Weber non fa loss augmentation anche se usa due loss, ma nel loro caso nella seconda loss non c'è la spiegazione.} 
%In our approach, we train an architecture from scratch by exploiting SHAP explanations to enhance performance.
\section{IMPACTX framework}
\label{sec:method}
\subsection{Adopted assumptions and notation}
In its simplest form, a typical ML classification system $\mathcal{A}$ can be usually viewed as the composition of two main components: a feature extractor $M$ (for example, in a feed forward DNN model, it usually corresponds to the DNN first layers) and a classification component $Q$ (usually corresponding, in a feed forward model, to the remaining part of the classification process) s.t. $\hat{y}=\mathcal{A}(\vec{x})=Q\big(M(\vec{x})\big) \in \{1,\dots K\}$ is the estimated class of the input $\vec{x}\in \mathbb{R}^d$. Without losing in generality, in this paper, we consider $Q$ as just the final function to compute the inferred class from a given vectors of $K$ scores, i.e. $Q(\vec{m})=\arg\max\big(\text{softmax}(\vec{m})\big)$ with $\vec{m}\in \mathbb{R}^K$. After a proper learning procedure, $M$ is a model able to represent a given $\vec{x}$ in a $K$-component array where the $k$-th component with $1\leq k \leq K$ represents a score of $\vec{x}$ to belong to the $k$-th class. An example of $M$ can be all the layers of a DNN before the final softmax function. However, in general every model able to project $\vec{x}$ in a given feature space can be adopted as $M$ (such as a sequence of layers of a DNN able to extract features from an input $\vec{x}$).
We define with $y \in \{1,\dots,K\}$ a scalar representing the correct label of an input $\vec{x}$  in a $K$-class classification problem. Let $S^T$ be a training dataset of $N$ labeled instances, i.e. $S^T = \{(\vec{x}^{(i)}, y^{(i)})\}_{i=1}^N$, and $S^E$ a test dataset composed of $J$ instances used only to evaluate an ML model, i.e. $S^E =\{ (\vec{x}^{(j)}, y^{(j)}) \}_{j=1}^J$. % our objective is to predict the correct class using a neural network $M$ trained on $S^T$.

\subsection{General description}
IMPACTX is a double-branch architecture such that, when applied to a classifier $\mathcal{A}$, the resulting architecture outperforms the standalone $\mathcal{A}(\vec{x})=Q\big(M(\vec{x})\big)$), both appropriately trained. In addition, this enhanced architecture  also provides input attribution maps relative to the output obtained.

\begin{figure*}[ht]
\centering
\includegraphics[width=1\textwidth]{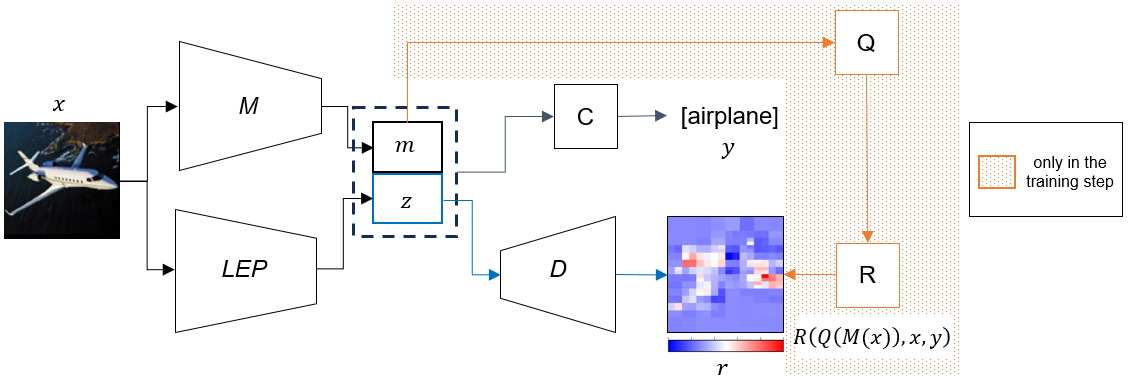} 
\caption{An overview of the IMPACTX framework. In the training phase of IMPACTX, both $M$ and $LEP$ receive $\vec{x}$, generating $\vec{m}$ and $\vec{z}$ respectively. These are combined for classification by $C$. In particular, $LEP$ and $D$ exploit the $R\big(\mathcal{A}(\vec{x})\big),\vec{x}, y)$ explanations. The architecture is trained using a loss function that merges MSE and CE to optimize explanation reconstruction and improve classification performance. In the inference step of IMPACTX, $C(\vec{m}, \vec{z})$ predicts the class $\hat{y}$ and $LEP-D(\vec{x})$ reconstructs the explanation $\vec{r}$ of the input $\vec{x}$. 
%In the training phase of IMPACTX, both $M$ and $LEP$ receive $\vec{x}^{(i)}$, generating $\vec{m}^{(i)}$ and $\vec{z}^{(i)}$ respectively. These are combined for classification by $C$. While the weights of $M$ remain frozen, $LEP$ and $D$ exploit the $\vec{r}^{(i)}$ explanations to enhance $M$'s classification. The architecture is trained using a loss function that merges MSE and CE to optimize explanation reconstruction and improve classification performance.
}
\label{fig:IMPACTX_arch}
\end{figure*}

IMPACTX framework is composed of two branches that interact each other (see figure \ref{fig:IMPACTX_arch}): 
\begin{enumerate}
    \item The first branch (on the top) is composed of a Feature Extractor $M$, able to extract significant features from the input with the goal to classify the input $\vec{x}$, and a $K$-class classifier $C$, able to return an estimated class $\hat{y}$ of the input $\vec{x}$. 
    \item The second branch (at the bottom)     \textcolor{black}{
     is responsible for the attention mechanism of IMPACTX. It}
 is composed of a Latent Explanation Predictor (LEP) module, able to extract essential information from the input features with the goal to compute an attribution map of the input respect to the classification response, and a Decoder $D$, able to effectively produce an attribution  map of the input $\vec{x}$. 
\end{enumerate}

Thus, the goal of IMPACTX is to build an estimated class $\hat{y}=C(\vec{m},\vec{z})$ exploiting both the $\vec{m}=M(\vec{x})$ and $\vec{z}=LEP(\vec{x})$ outputs, and at the same time to obtain a predicted attribution map $\hat{\vec{r}}$ with respect to $\hat{y}$. 

%As we discuss in detail in Section \ref{sec:training}, during the inference phase the top-branch is influenced by the bottom branch, while the bottom branch is used to the during the training phase the bottom branch is influenced 
In the next section we will discuss in detail how IMPACTX can be trained to obtain this goal.

%Summarizing, the IMPACTX training framework comprises several key components:  (i) the Latent Explanation Predictor $LEP$ and a Decoder $D$ are trained to extract significant hidden information $\vec{z}^{(i)}$ related to the explanations and reconstructing the original attribution map  $\vec{r}^{(i)}$ from its encoding $\vec{z}^{(i)}$, and (ii) the final classifier $C$ designed to exploit $\vec{m}^{(i)}$ and $\vec{z}^{(i)}$ to predict the effective class $y^{(i)}$. Both $M$ and $LEP$ receive the input $\vec{x}^{(i)}$. The resulting outputs, $M(\vec{x}^{(i)})$ and $LEP(\vec{x}^{(i)})$, are concatenated and passed to $C$.

%\input{_alg}

\subsection{Training IMPACTX}
\label{sec:training}
The training phase of the IMPACTX approach is depicted in figure \ref{fig:IMPACTX_arch}.
%\textcolor{red}{ and outlined in the pseudocode presented in Algorithm  \ref{algo:method_main}}
Both $M$ and $LEP$ receive $\vec{x}$ as input, producing the corresponding outputs $\vec{m}$ and $\vec{z}$. These outputs are concatenated and forwarded to the classifier $C$. Additionally, $\vec{z}$ is decoded by the decoder $D$, responsible for reconstructing the explanation $\vec{r}$. In other words, $LEP$ and $D$ act as an Encoder-Decoder which is constrained to learn an encoding of the explanations by the internal variables $\vec{z}$. $C$ leverages the combined knowledge of $M$  and $LEP$. Therefore, for a new data point $\vec{x}^{(j)}$, the  estimated output $\hat{y}^{(j)}$ is defined as: $$\hat{y}^{(j)} = \arg\max\bigg( \text{softmax} \Big(C\big(M(\vec{x}^{(j)}), LEP(\vec{x}^{(j)})\big)\Big)\bigg).$$ %Assuming that $M$ can be any pre-trained neural network, learning the proposed framework requires the training of Predictor $P$, Decoder $D$, and classifier $C$ without utilizing unlabeled data $S^E$.
\\Consequently, IMPACTX wants to solve the classification task and, at the same time, to construct a predicted attribution map $\vec{r}$ using the decoder $D$ and the $LEP$ module. 

Importantly, the effective IMPACTX training can be made in at least two ways:

- \textit{Single-stage training}: All IMPACTX modules are trained simultaneously on $S^T$, with the attribution maps $\vec{r}^{(i)}$ for each sample in $S^T$ generated at the end of each training iteration. \textcolor{black}{In this case, classification performance can be improved by using increasingly accurate attribution maps, $\vec{r}^{(i)}$. Ideally, this creates a positive feedback loop where both the classification and the generation of attribution maps are mutually enhanced. Conversely, computing the attribution maps at each iteration can be computationally expensive and, at the same time, the generated attribution maps $\vec{r}^{(i)}$ may have very poor significance in the early learning epochs since $M$, $LEP$ and $D$ weights are initialised to random values. %This situation could also lead to the worst results in the $C$ classifier and in some cases, without special precautions, also in the last learning epochs. 
%Instead, the advantage of this strategy is obviously a single-stage training for all IMPACTX modules. In the optimal case, there's the possibility to improve the classification performance by using better and better attribution maps $\vec{r}^{(i)}$ and in an ideal loop, this enhances both the classification and the generation of attribution maps.
}

- \textit{Two-stage training}:  This training approach is divided into two stages. In the first stage, the parameters of the whole classifier  $\mathcal{A}(\cdot)$ are trained and initially evaluated on $S^T$. Then, at the end of the first training stage, the attribution maps $\vec{r}^{(i)}$ for each sample in $S^T$ are produced with respect the true class label. In the second training stage, the remaining modules $C$, $D$, and $LEP$ are trained while keeping $M$ frozen. In particular, the targets of the branch $LEP$-$D$ are the corresponding attribution maps previously computed at the end of the first stage.  Since the attribution maps are computed just one time, two-stage training results less expensive than single-stage training. 
%\textcolor{red}{The disadvantage of this approach is that only one generation of attribution maps $\vec{r}^{(i)}$ is used. This could limit the performance gain.}\textcolor{blue}{Andrea: ???} \sal{vorrei dire, che rispetto all'altro tipo di addestramento, qui vengono generate una sola volta le spiegazioni e per questo motivo (nel caso ideale) l'incremento di performance potrebbe essere inferiore. Se non va bene, per ogni finalità, si può anche togliere}

In both the training approaches, the architecture is trained using a loss function that combines together Mean Squared Error (MSE) between the true class attribution map $\vec{r}^{(i)}$ (see sec. \ref{sec:method_genR}) and the output of the $LEP$-$D$ branch, and the Cross Entropy (CE) loss between the true class label $y^{(i)}$ and the prediction from $C$. The resulting loss function can be formalized as follows: 
\begin{equation}
    L = CE\big(y^{(i)}, C(\vec{m}^{(i)}, \vec{z}^{(i)})\big) + \lambda \cdot MSE\big(\vec{r}^{(i)}, LEP-D(\vec{x}^{(i)})\big)
\label{eq:loss}
\end{equation} where $\lambda$ represents a regularization parameter. This approach lead $\vec{z}^{(i)}$ to be optimized respect to the attribution reconstruction error while maintaining robust classification performance simultaneously.

\subsection{Generating the attribution-based explanations} 
\label{sec:method_genR}
We use an XAI attribution method $R$ to generate attribution maps $\vec{r}$ on $\vec{x}$ about the true class label $y$ based on $\mathcal{A}(\vec{x})$. In particular, for each available training data $\vec{x}^{(i)}$ in $S^T$, an attribution map $\vec{r}^{(i)}$ corresponding to the true class label $y^{(i)}$ when $\vec{x}^{(i)}$ is the input of $\mathcal{A}(\vec{x}^{(i)})=Q\big(M(\vec{x}^{(i)})\big)$, is produced adopting $R$. 
The objective is to get an attribution map aligned with the true class label $y^{(i)}$ for each training data $\vec{x}^{(i)}$. Therefore, we adopt as $R$ an XAI method that provides explanations not only for the predicted class, but also for each possible class, which we denote as $R\big(\mathcal{A}, \vec{x}, k \big)=\vec{r}_k$, where $k$ represents the class for which we need the explanation. Therefore, the attribution map corresponding to the true class label provided in the training data results $\vec{r}^{(i)}=R\big(\mathcal{A}(\vec{x}^{(i)}), \vec{x}^{(i)}, y^{(i)}\big)$.
\section{Experimental setup}
\label{sec:exp}
%In this section, we present a comprehensive experimental evaluation of the proposed IMPACTX framework. This series of experiments has a dual goal: to evaluate IMPACT's ability to improve model performance, and to provide meaningful explanations in the form of attribution maps. 

In this section, we present a series of experiments aimed at evaluating the IMPACTX framework with respect to two different capabilities:
to improve the performance of ML models and to provide meaningful explanations in the form of attribution maps.
%Our assessment begins with a performance evaluation of IMPACTX on several classification tasks, 
%IMPACTX performance evaluation is obtained  on several standard classification tasks, comparing its results with similar methods in the literature, namely ABN \citep{} and WAE \citep{}, and with IMPACTX itself without the attention mechanism providing by the bottom branch (we considered it as \textit{baseline}). 
The performance of IMPACTX is evaluated on several standard classification tasks, 
comparing its results with similar methods in the literature, namely ABN \citep{fukui2019attention} and WAE \citep{apicella2023shap}, and with IMPACTX itself without the attention mechanism provided by the bottom branch (we considered it as \textit{baseline}).
Following this, we present both qualitative and quantitative analyses to assess the validity of the attribution maps generated by IMPACTX. %This dual approach assesses the IMPACTX's ability to not only enhance model performance but also to provide meaningful explanations in form of attribution maps. 
The attribution maps' qualitative analysis focuses on how the explanations highlight significant parts of the elements to be classified, which we expect to be aligned with human user expectations, providing intuitive insights into the model's decision-making process. Meanwhile, the quantitative evaluation employs MoRF (Most Relevant First) \citep{samek2016evaluating} curves to measure the relevance of the features identified by the attribution maps, therefore if the provided attribution maps can be considered reliable explanations of the framework. 

%\roberto{Questa sezione ora bisogna cambiarla  pensarla,forse, un po' meglio. Cosi' come la parte precedente sul learning. Discutiamone. inoltre bisogna dire che usiamo anche ABN e WAE}
%\sal{meglio come?}

%%\begin{enumerate}%\item \textit{Model without IMPACTX:} training of a model with the same architecture $M$  model on the training dataset $S^T=\{\vec{x}^{(i)}, y^{(i)}\}_{i=1}^{n}$ and computing its accuracy on the test set $S^E=\{\vec{x}^{(j)}, y^{(j)}\}_{j=1}^{m}$, where the labels $y^{(j)}$ are used exclusively for evaluation.
%\item \textit{Calculation of attribution:} we derive the attribution maps $\{\vec{r}^{(i)}\}_{i=1}^{n}$ using the model $M$ with each instance $\textbf{x}^{(i)}$ in the training set $S^T$, with respect to their true class $y^{(i)}$, as outlined in Sec. \ref{sec:method_genR}.
%\item \textit{IMPACTX Training:} learning the parameters of the modules $LEP$, $D$, and $C$ (see Fig. \ref{fig:IMPACTX_arch} and Sec. \ref{sec:trainPhase}) using the training data $S^T$.  In this step we derive the attribution maps $\{\vec{r}^{(i)}\}_{i=1}^{n}$ using the model $M$ with each instance $\textbf{x}^{(i)}$ in the training set $S^T$, with respect to their true class $y^{(i)}$, as outlined in Sec. \ref{sec:method_genR}.
%The pre-trained parameters of $M$ are not changed during this phase.
%\item \textit{Model Evaluation:} using the outputs of the Latent Explanation Predictor $LEP$ and the model $M$ as input of $C$ to predict the labels of the unseen data $S^E$ and compare the performance with that of the baseline on the same data set.
%\end{enumerate}
The following subsections provide detailed information on the datasets used, the attribution generator algorithm $R$, the IMPACTX modules' architectures and the learning procedure.

\subsection{Datasets}
The benchmark datasets used in this study include CIFAR-10, CIFAR-100, and STL-10.
\begin{itemize}
    \item CIFAR-10 \citep{Krizhevsky09} comprises 60,000 color images categorized into ten classes: airplane, automobile, bird, cat, deer, dog, frog, horse, ship, and truck. The dataset is split into 50,000 training images and 10,000 test images, all sized $32 \times 32$ pixels.
    \item CIFAR-100 \citep{Krizhevsky09} contains 60,000 color images grouped into 100 categories, with 50,000 training images and 10,000 test images, all sized $32 \times 32$ pixels.
    \item STL-10 \citep{coates2011analysis} dataset comprises images classified into ten classes: airplane, bird, car, cat, deer, dog, horse, monkey, ship, and truck. Each image is $96 \times 96$ pixels, and the dataset includes 5,000 training images and 8,000 test images.
\end{itemize}

\subsection{Explanation Generator Algorithm}
%Since our goal is to obtain attribution maps aligned with the true class label, we adopt the XAI SHAP method \citep{NIPS2017_7062} as $R$. SHAP, a prominent technique in XAI, provides insights into the contribution of each feature to a model's predictions. In particular, SHAP provides explanations not only for the predicted class, but also for each possible class. 
We have adopted the SHAP method \citep{NIPS2017_7062} as $R$. This choice was motivated by the fact that SHAP is a  prominent technique in XAI that provides explanations in terms of attribution maps, and provides explanations not only for the predicted class, but also for each possible class, as required by our approach. 
For this study, in particular, we employed the \textit{Partition Explainer algorithm} as specific version of SHAP, with $2000$ evaluations to obtain the final explanations \citep{NIPS2017_7062}. 
%The explanations have been normalized between 0 and 1 using the minimum and maximum values computed on the training set.

\subsection{IMPACTX modules and the two-stage training}
\label{subsec:AdoptedModelsTraining}
We evaluated IMPACTX using three distinct models for $M$: LeNet-5 \citep{lecun1998gradient}, MobileNet \citep{howard2017mobilenets} pre-trained on ImageNet \citep{deng2009imagenet}, and  EfficientNet-B2 \citep{tan2019efficientnet} pre-trained on the Noisy Student weights \citep{xie2020self}. 
The focus of this study is oriented
to the interpretation of the model and to the explanation
of the influence of the input characteristics on the models’
prediction. %\begin{itemize}   \item LeNet-5 \citep{lecun1998gradient}.\item  MobileNet \citep{howard2017mobilenets} pre-trained on  a ImageNet weights \citep{deng2009imagenet}.\item EfficientNet-B2 \citep{tan2019efficientnet} pre-trained on the Noisy Student weights \citep{xie2020self}.\end{itemize}

\begin{figure}[ht]
    \centering
\includegraphics[width=0.8\textwidth]{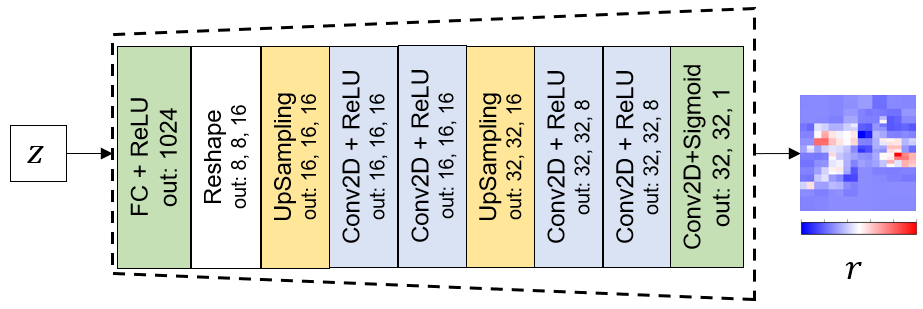}
    \caption{The Decoder architecture designed for the CIFAR-10 and CIFAR-100 datasets. The architecture is composed of convolutional (Conv2D), fully connected (FC), and UpSampling layers. The kernel size is $3 \times 3$ for all the convolutional layers, while the number of filters is given by the third dimension of the output shape.}
    \label{fig:Decoder}
\end{figure}

The Latent Explanation Predictor $LEP$ module is built starting from the architecture of selected $\mathcal{A}$, replacing the last fully-connected layers by a fully-connected layer with a dimensionality of $512$
%\textcolor{red}{, i.e. the coding dimension of the input image,}\textcolor{blue}{Andrea: ???} \sal{la dimensione della codifica, non lo diciamo da nessuna parte in modo esplicito, non va bene?}\textcolor{blue}{Andrea: ok, ma non si capisce codifica di che...} \sal{corretto, vedi se "suona" bene} 
and a sigmoid activation function. The architecture of the Decoder used for CIFAR-10 and CIFAR-100 datasets is depicted in fig. \ref{fig:Decoder}. In case of STL-10, an additional sequence of [Conv, Conv, UpSampling] layers was included in the decoder due to the different input dimensions of STL-10 images compared to CIFAR-10/100. It should be emphasised again that the Latent Explanation Predictor - Decoder ($LEP-D$) extracts insights and relationships concerning both the visual features of the input image and the corresponding attribution scores found during the training phase. The outputs of $M$ and $LEP$, representing the extracted features from $\vec{x}$ and the encoding attribution $\vec{z}$ respectively, are concatenated and then fed into the final classifier $C$. The classifier $C$ is a shallow fully-connected neural network with outputs equal to the number of classes. 

\begin{table}[ht]
\centering
\caption{Variation ranges for the grid search optimization strategy.}
\scalebox{0.8}{

\begin{tabular}{lll}
\hline
\multicolumn{1}{c}{} & \multicolumn{1}{c}{\textbf{Hyperparameter}} & \multicolumn{1}{c}{\textbf{Range}} \\ \hline
\multirow{3}{*}{$\mathcal{A}$} & Batch Size & \{16, 32, 64\} \\
 & Learning Rate & {[}0.0001, 0.01{]} with step of 0.0005 \\
 & Validation Fraction & \{0.05, 0.1, 0.2\} \\ \hline
$IMPACTX$ \hspace{0.5cm} & $\lambda$ & {[}0.2, 1.4{]} with step of 0.2 \\ \hline
\end{tabular}

}

\label{tab:tab_hp}
\end{table} 

As discussed in sec. \ref{sec:training}, we used the two-stage training because it is less expensive than single-stage training. 
Thus, following this learning strategy, in the first stage learning, the hyperparameters batch size, learning rate, and validation fraction values were determined through a grid-search approach, and each $\mathcal{A}$ model is fine-tuned on each selected dataset. Grid-search ranges are detailed in \cref{tab:tab_hp}. After the first stage learning, 
in the second stage learning, the parameters of $LEP$, $D$, and $C$ are learned while maintaining the parameters of $M$ fixed (see fig. \ref{fig:IMPACTX_arch}).
%the best final models are considered as frozen models $M$ on which the second stage training is applied. 
The $D$ architecture is reported in fig. \ref{fig:Decoder}. The optimal value for the $\lambda$ regularization parameter in eq. \ref{eq:loss} was established via hyperparameter grid-search, and the values used for the search are detailed in \cref{tab:tab_hp}.
Final performance on the test set for each dataset is computed. 

%Subsequently, SHAP attribution maps are generated for each input within the training data, taking into account their true class labels.

\subsection{Evaluation measures}
%In this study, to evaluate IMPACTX we conducted our experiments using the two-stage training approach described in sec. \ref{sec:training}. 
%This choice was primarily driven by computational resource constraints. 
We evaluated IMPACTX for its performance in the classification tasks and for the attribution maps that it generates. Performances on classification tasks were measured in terms of accuracy on the standard test sets provided with each dataset. The proposed method was subjected to a comparative analysis with the baseline $\mathcal{A}$ (which can be seen as IMPACTX itself without the attention mechanism provided by the bottom branch), ABN \citep{fukui2019attention} and WAE \citep{apicella2023shap} approaches.
Instead, the attribution maps are evaluated both qualitatively and quantitatively, reporting respectively examples of the obtained attribution maps and MoRF (Most Relevant First) curves \citep{samek2016evaluating}.  MoRF curves are widely used in XAI research to assess proposed explanations  \citep{kakogeorgiou2021evaluating, apicella2022exploiting, tjoa2022quantifying, islam2022systematic}. Essentially, image regions are iteratively replaced by random noise and fed to the ML model, following the descending order of relevance values indicated by the attribution map. Thus, the more relevant the identified features are for the classification output, the steeper the curve. MoRF curves offer a quantitative measure of how well an attribution map explains the behavior of an ML system. Summarizing, the qualitative aspect is important for understanding how closely the obtained explanation aligns with the user's intuitive expectations, while the quantitative evaluation aims to provide a measure of how valid the attribution maps are as explanations. %Finally, IMPACTX explanations obtained by the $LEP-D$ branch are compared with explanations obtained by the SHAP method applied to the $Q\big(M(\cdot)\big)$.

%In the IMPACTX framework, the hyperparameter grid search was omitted since the $LEP$ predictor and the $M$ baseline share identical architectures. As a result, the optimal hyperparameter values for the $LEP$, $D$, and $C$ modules in IMPACTX correspond to those discovered for the $M$ baseline module through grid search, as delineated in \cref{tab:tab_hp}. Moreover, a $\lambda$ value of 1 was selected to ensure equal weighting between the two losses CE and MSE.
%\andrea{sta parte la ometterei. Inoltre, metterei che lambda è stato usato come hyperparametro con valori inventati} \sal{credo sia una cosa positiva dire che non è stata fatta anche su impactx una grid search, o a limite si puà scrivere per question di tempi computazioniali non è stata fatta. Lambda con valori inventati?}\andrea{aggiungi nei parametri della grid search che hai fatto una ricerca pure su lambda inventandoti i range} \sal{non mi convince e non è quello che è stato fatto}\andrea{lo so che non è quelllo che è stato fatto, ma non voglio rischiare che un revisore caghi il cazzo su sto punto che, dati i risultati decenti già adesso, oggettivamente sarebbe una stronzata (per intenderci, non stiamo imbrogliando presentando come migliori di quello che sono, ma peggiori di quello che potrebbero essere, e non voglio che il revisore per fare lo splendido ce lo bocci o vada in major dicendo "cosa succede con vari valori di labmda?"} \sal{ok}

\section{Results and Discussion}
\label{sec:res}

\subsection{Performance}
Tab. \ref{tab:tab_results} shows the performance of different architectures $\mathcal{A}(\cdot)$ (that are LeNet-5, MobileNet, and EfficientNet-B2) on CIFAR-10, CIFAR-100, and STL-10 test sets with (IMPACTX column) and without (baseline column) IMPACTX, in terms of accuracy. In the same table, results obtained by ABN and WAE are reported. 

\begin{table}[ht]
\centering
\caption{Accuracy (\%) scores on CIFAR-10, CIFAR-100 and STL-10 test sets. Best results are reported in in boldface}
\scalebox{0.8}{

\begin{tabular}{cccccc}
\hline
 & \textbf{Model} & \textbf{Baseline $\mathcal{A}$} & \textbf{IMPACTX}  & \textbf{ABN} \citep{fukui2019attention} & \textbf{WAE} \citep{apicella2023shap} \\ \hline
\multirow{3}{*}{\textbf{CIFAR-10}} & LeNet-5 & 67.96 & \textbf{70.65} & 68.49 & 68.10 \\ \cline{2-6} 
 & MobileNet & 94.63 & \textbf{96.27} & 95.41 & 94.75 \\ \cline{2-6} 
 & EfficientNet-B2 & 98.06 & \textbf{98.38} & 98.19 & 98.08\\ \hline
\multirow{3}{*}{\textbf{CIFAR-100}} & LeNet-5 & 36.23 & \textbf{39.49} & 37.62 & 36.20\\ \cline{2-6} 
 & MobileNet & 76.72 & \textbf{79.37} & 78.24 & 76.98 \\ \cline{2-6} 
 & EfficientNet-B2 & 87.88 & \textbf{89.06} & 88.89 & 88.27\\ \hline
\multirow{3}{*}{\textbf{STL-10}} & LeNet-5 & 52.59 & \textbf{55.86} & 53.36 & 52.99 \\ \cline{2-6} 
 & MobileNet & 91.07 & \textbf{95.79} & 92.71 & 91.42\\ \cline{2-6} 
 & EfficientNet-B2 & 98.76 & \textbf{99.00} & 98.59 & 98.65 \\ \hline
\end{tabular}

}
\label{tab:tab_results}
\end{table}

One can note that IMPACTX produced an uniform improvements with respect the baseline on all the selected datasets independently from the classification architecture used. At same way, IMPACTX obtained better results with respect the other selected approaches.

%an improvement in accuracy from $67.96 \%$ (baseline) to $70.65 \%$ using LENET-5 as architecture. Similarly, with MobileNet the accuracy increased from $94.63 \%$ to $96.27 \%$. When EfficientNet-B2 is used as classification's architecture, our approach maintains a similar but slightly higher accuracy ($98.38 \%$) compared to the baselines ($98.06 \%$). Similar results are 

%With LeNet-5 as baseline model $M$, the accuracy increased from $36.23\%$ (baseline) to $39.49 \%$ using the IMPACTX approach. Using MobileNet as baseline model led to a more significant improvement, with accuracy climbing from $76.72 \%$ to $79.37 \%$. Instead, the improvement observed with EfficientNet-B2 was more moderate, increasing from an accuracy of $87.88 \%$ to $89.06 \%$.

%\subsubsection{Results on STL-10}
%Using LeNet-5 as baseline model $M$, the proposal's accuracy on STL-10 increased from $52.59 \%$ to $55.86 \%$, showing a significant enhancement in accuracy compared to the baseline. Similarly, with MobileNet exhibited an improvement from $91.07 \%$ to $95.79 \%$. Conversely, when considering EfficientNet-B2 as baseline model $M$, the proposal maintained a higher accuracy ($99.00 \%$), albeit comparable with the baseline ($98.76 \%$).

\begin{figure}[h!]
\centering
    \centering
    \includegraphics[width=1\textwidth]{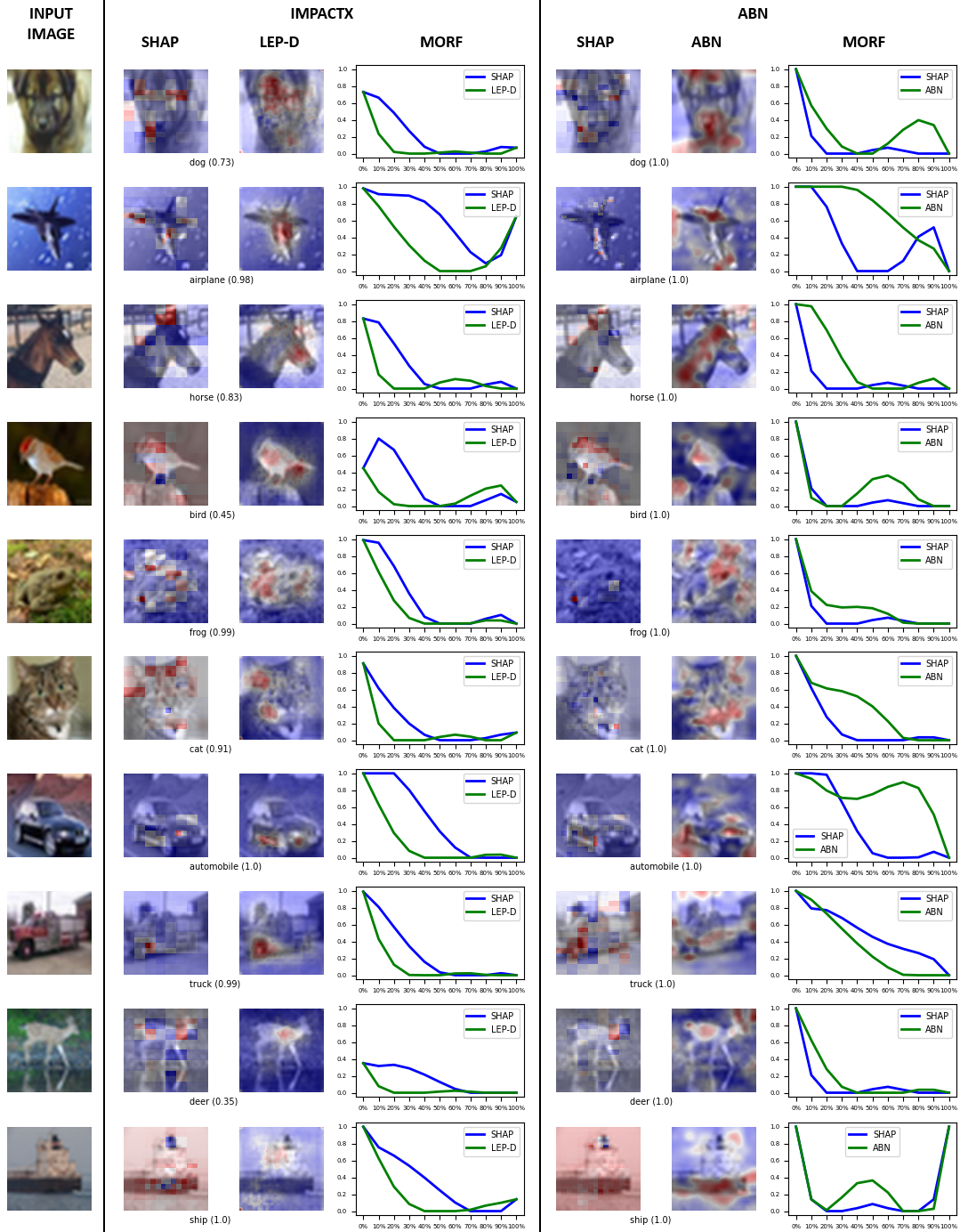}
\caption{Images from the CIFAR-10 test set. The images have been filtered for better visualisation.
} 
\label{images_CIFAR10}
\end{figure}

\begin{figure}[h!]
\centering
    \centering
    \includegraphics[width=1\textwidth]{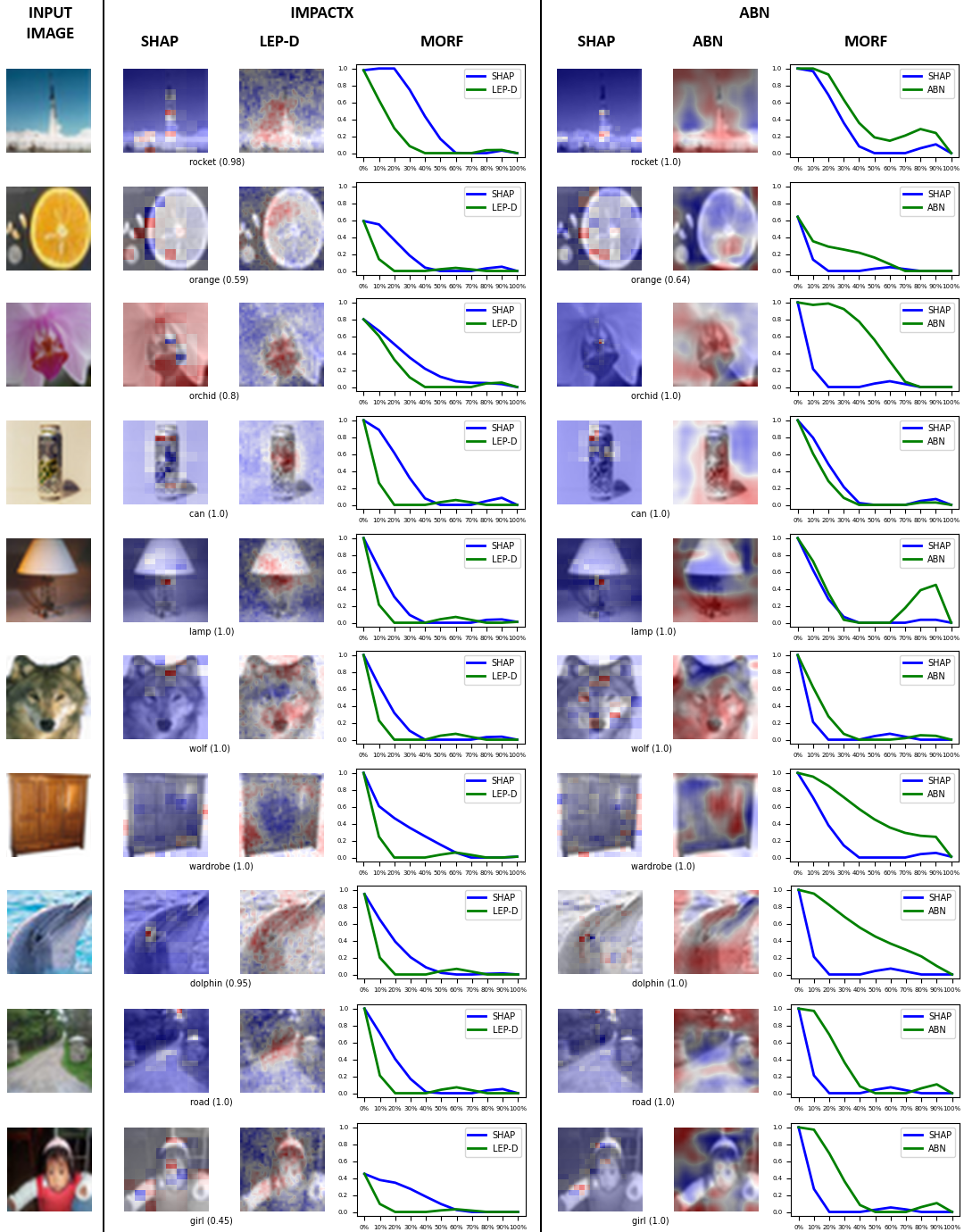}
\caption{Images from the CIFAR-100 test set. The images have been filtered for better visualisation.} 
\label{images_CIFAR100}
\end{figure}

\begin{figure}[h!]
\centering
    \centering
    \includegraphics[width=1\textwidth]{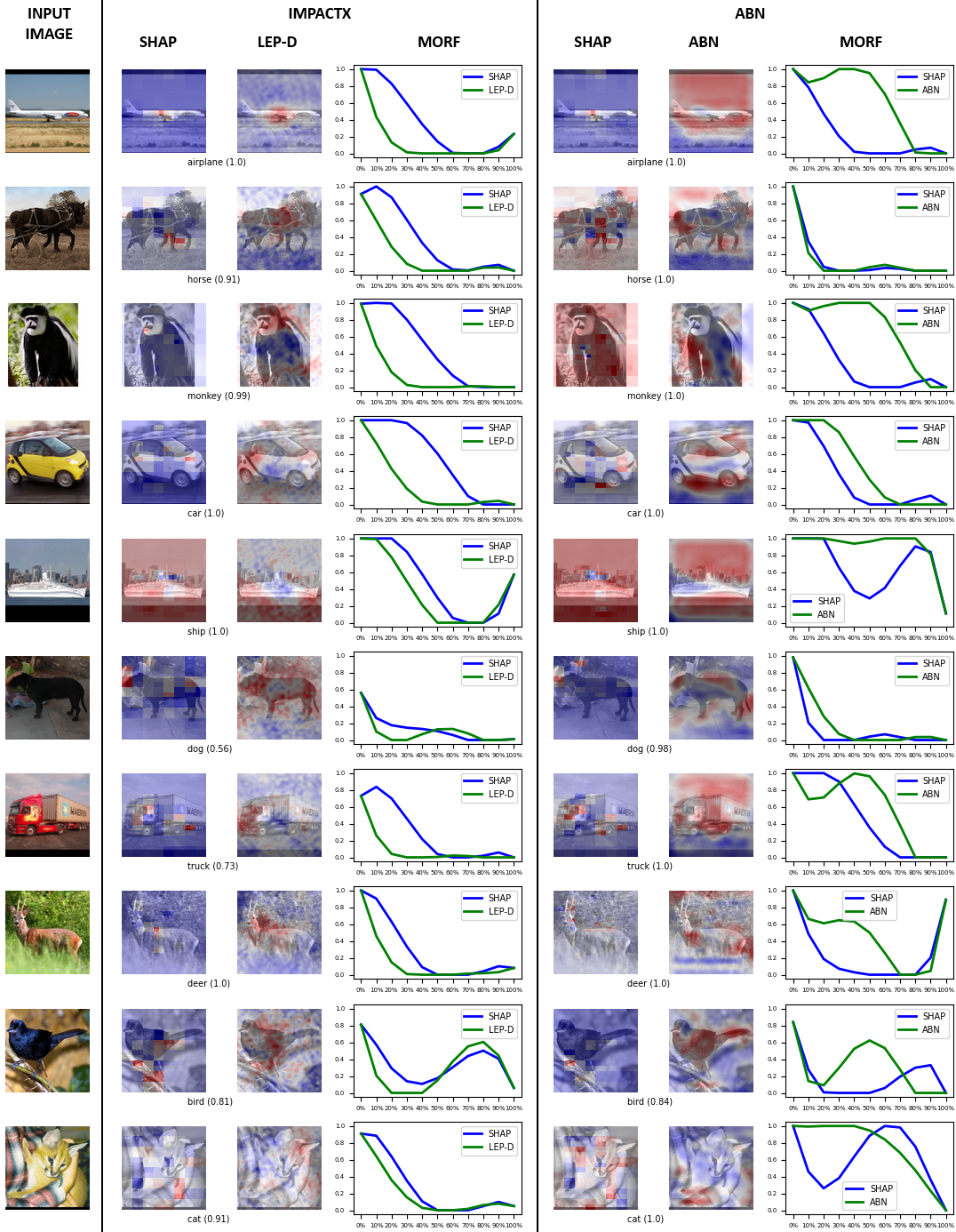}
\caption{Images from the STL-10 test set. The images have been filtered for better visualisation.} 
\label{images_STL10} 
\end{figure}

\subsection{Evaluating attribution maps}
In this section we want to evaluate if the attribution maps directly obtained by IMPACTX can be considered as explanations of the IMPACTX classification responses. To this aim, we compare them with the explanations given by $R$ (that is, in this experimental setup, SHAP) outputs when applied on IMPACTX itself.  
In figures \ref{images_CIFAR10}, \ref{images_CIFAR100} and \ref{images_STL10}
(left side), examples from the CIFAR-10, CIFAR-100, and STL-10 test sets are presented (column 1). The examples are reported considering the experiments made on \textit{LeNet-5}. For each example, the class and score given by IMPACTX are provided. Furthermore, in the same figures are reported the attribution maps obtained by the explanation method SHAP (column 2), along with the attribution maps  obtained by IMPACTX (column 3). For each input, MoRF curves for SHAP and IMPACTX attribution maps are also reported (column 4). The same comparison is  reported for ABN architecture (right side). Interestingly, while both IMPACTX and ABN qualitatively emphasize input features that intuitively seem more relevant for their respective classes, the MoRF curves reveal different behaviors quantitatively. Specifically, the MoRF curves indicate that the most relevant elements identified by ABN have a lesser impact on classification compared to those selected by SHAP. Conversely, an opposite pattern is observed between IMPACTX and SHAP. The most relevant features highlighted by IMPACTX prove to be more influential for classification than those identified by SHAP. %It is possible to note that, in the majority of cases, the IMPACTX attribution maps focus on input regions intuitively more relevant from the human point of view to the true output compared to the explanation corresponding to the $Q\big(M(\cdot)\big)$ response. %Thus, highlighting the effectiveness of \textit{IMPACTX} approach in generating knowledge useful for the correct prediction.

\subsection{Discussion}
\label{sec:discussion}
%The experimental results show that IMPACTX delivers higher performance in the experimented classification tasks compared to ML models not adopting IMPACTX framework, while also providing attribution maps that can be considered more reliable explanations than those offered by post-hoc XAI methods.
The experimental results show that IMPACTX performs better on the classification tasks than ML models that do not use the IMPACTX framework, while providing attribution maps that can be considered more reliable explanations than those provided by post-hoc XAI methods.
In particular, on the performance side, the interaction between the IMPACTX branches enhances the model's ability to focus on relevant features, thereby improving classification performance. During training, IMPACTX emphasizes features identified as relevant to the model's decisions, leading to more accurate outcomes. %Indeed, the training process of IMPACTX is aimed to focus on features identified as important to the model's decisions.
In fact, IMPACTX's training process is designed to focus on the features identified as important to the model's decisions.
Furthermore, regarding the construction of the attribution maps by the bottom branch, the ability of IMPACTX in providing accurate attribution maps can be attributed to its integrated approach of decision and justification at inference time, which contrasts with the external nature of post-hoc methods such as SHAP. This integration allows IMPACTX to generate attribution maps that more accurately reflect the model's decision-making process. Interestingly, attribution maps generated by IMPACTX seems to be better explanations compared to ABN. This is likely because IMPACTX's attentional mechanism is designed to consider the true class label during training, whereas ABN's CAM-based approach relies on class scores obtained during training.

%Finally, the fact that the attribution maps qualitatively seem to match user expectations, together with their reliability in explaining model's decisions confirmed by MoRF curves, indicates that the ML system tends to focus on aspects similar to human attention to make its predictions. 

%This can be attributed to the fact that the most informative features of the input data identified by both humans and IMPACTX are the same and that are effectively discriminative for the task at hand.

%\input{_discussion}
\section{Conclusion}
\label{sec:conclusion}

In conclusion, experimental results showed that IMPACTX is able to include XAI strategies into the ML pipeline positevely affecting the performance. Furthermore, IMPACTX computes valid explanations for its outputs at inference time, without relying on external XAI methods after the inference process. In particular, the experimental assessment shows the  effectiveness of the incorporated latent explanation predictor $LEP$ in extracting useful knowledge. This knowledge can be used both to the ML task and to provide explanations about the model's outputs. It's noteworthy that $LEP$ is constructed based on explanations related to true classes extracted by a specific XAI method from the training set where the model was trained, while the decoder $D$ is able to provide useful explanations regarding the model outputs. As future research, it could be valuable to explore the impact of this approach by employing different XAI methods, architectures, and datasets than those used to train the IMPACTX framework.  Furthermore, in this study IMPACTX was trained using the two-stage training approach described in sec. \ref{sec:training}, and this choice was primarily driven by computational resource constraints. Thus, it may be interesting in future work to compare two-stage training with single-stage training. 
Finally, we note that the qualitative analysis of the IMPACTX attribution maps seems to indicate that the ML system, together with its reliability in explaining the model decisions confirmed by the MoRF curves, tends to focus on aspects similar to those that could be obtained by human attention when making decisions. 
Although this aspect should be better investigated in future work, this possible alignment suggests that the model has effectively learned to prioritise features that are inherently relevant to the task, similar to how a human would approach the same task. Notably, it seems that such a human-like focus not only enhances the interpretability of the model's decisions, but also contributes to improved performance in performing the ML tasks, as shown in sec. \ref{sec:res}. 

%\bibliographystyle{plainnat}
\bibliographystyle{plain}
 \bibliography{b}

@inproceedings{ijcai2017p371,
  author    = {Andrew Slavin Ross and Michael C. Hughes and Finale
Doshi-Velez},
  title     = {Right for the Right Reasons: Training Differentiable
Models by Constraining their Explanations},
  booktitle = {Proceedings of the Twenty-Sixth International Joint
Conference on
               Artificial Intelligence, {IJCAI-17}},
  pages     = {2662--2670},
  year      = {2017},
  doi       = {10.24963/ijcai.2017/371},
  url       = {https://doi.org/10.24963/ijcai.2017/371},
}

@article{apicella2019explaining,
	author = {Apicella, A. and Isgrò, F. and Prevete, R. and Sorrentino, A. and Tamburrini, G.},
	title={Explaining classification systems using sparse dictionaries},
	year={2019},
	journal={ESANN 2019 - Proceedings, 27th European Symposium on Artificial Neural Networks, Computational Intelligence and Machine Learning},
	pages = {495 – 500},
}

@inproceedings{ribeiro2016should,
  title={"Why should i trust you?" Explaining the predictions of any classifier},
  author={Ribeiro, Marco Tulio and Singh, Sameer and Guestrin, Carlos},
  booktitle={Proceedings of the 22nd ACM SIGKDD international conference on knowledge discovery and data mining},
  pages={1135--1144},
  year={2016}
}

@article{annuzzi2023impact,
  title={Impact of nutritional factors in blood glucose prediction in type 1 diabetes through machine learning},
  author={Annuzzi, Giovanni and Apicella, Andrea and Arpaia, Pasquale and Bozzetto, Lutgarda and Criscuolo, Sabatina and De Benedetto, Egidio and Pesola, Marisa and Prevete, Roberto and Vallefuoco, Ersilia},
  journal={IEEE Access},
  volume={11},
  pages={17104--17115},
  year={2023},
  publisher={IEEE}
}

@article{schramowski2020making,
  title={Making deep neural networks right for the right scientific reasons by interacting with their explanations},
  author={Schramowski, Patrick and Stammer, Wolfgang and Teso, Stefano and Brugger, Anna and Herbert, Franziska and Shao, Xiaoting and Luigs, Hans-Georg and Mahlein, Anne-Katrin and Kersting, Kristian},
  journal={Nature Machine Intelligence},
  volume={2},
  number={8},
  pages={476--486},
  year={2020},
  publisher={Nature Publishing Group UK London}
}

@article{schoonderwoerd2021human,
  title={Human-centered XAI: Developing design patterns for explanations of clinical decision support systems},
  author={Schoonderwoerd, Tjeerd AJ and Jorritsma, Wiard and Neerincx, Mark A and Van Den Bosch, Karel},
  journal={International Journal of Human-Computer Studies},
  volume={154},
  pages={102684},
  year={2021},
  publisher={Elsevier}
}

@article{apicella2022exploiting,
  title={Exploiting auto-encoders and segmentation methods for middle-level explanations of image classification systems},
  author={Apicella, Andrea and Giugliano, Salvatore and Isgr{\`o}, Francesco and Prevete, Roberto},
  journal={Knowledge-Based Systems},
  volume={255},
  pages={109725},
  year={2022},
  publisher={Elsevier}
}

@article{samek2016evaluating,
  title={Evaluating the visualization of what a deep neural network has learned},
  author={Samek, Wojciech and Binder, Alexander and Montavon, Gr{\'e}goire and Lapuschkin, Sebastian and M{\"u}ller, Klaus-Robert},
  journal={IEEE transactions on neural networks and learning systems},
  volume={28},
  number={11},
  pages={2660--2673},
  year={2016},
  publisher={IEEE}
}

@article{miller2019explanation,
  title={Explanation in artificial intelligence: Insights from the social sciences},
  author={Miller, Tim},
  journal={Artificial intelligence},
  volume={267},
  pages={1--38},
  year={2019},
  publisher={Elsevier}
}

@article{lipton2018mythos,
  title={The Mythos of Model Interpretability: In machine learning, the concept of interpretability is both important and slippery.},
  author={Lipton, Zachary C},
  journal={Queue},
  volume={16},
  number={3},
  pages={31--57},
  year={2018},
  publisher={ACM New York, NY, USA}
}

@article{arrieta2019explainable,
  title={Explainable Artificial Intelligence (XAI): Concepts, Taxonomies, Opportunities and Challenges toward Responsible AI},
  author={Arrieta, Alejandro Barredo and D{\'\i}az-Rodr{\'\i}guez, Natalia and Del Ser, Javier and Bennetot, Adrien and Tabik, Siham and Barbado, Alberto and Garc{\'\i}a, Salvador and Gil-L{\'o}pez, Sergio and Molina, Daniel and Benjamins, Richard and others},
  journal={Information Fusion},
  volume={58},
  issue={June},
  pages={82-115},
  year={2020},
  publisher={Elsevier}
}

@article{weber2022beyond,
  title={Beyond explaining: Opportunities and challenges of XAI-based model improvement},
  author={Weber, Leander and Lapuschkin, Sebastian and Binder, Alexander and Samek, Wojciech},
  journal={Information Fusion},
  year={2022},
  publisher={Elsevier}
}

@article{yeom2021pruning,
  title={Pruning by explaining: A novel criterion for deep neural network pruning},
  author={Yeom, Seul-Ki and Seegerer, Philipp and Lapuschkin, Sebastian and Binder, Alexander and Wiedemann, Simon and M{\"u}ller, Klaus-Robert and Samek, Wojciech},
  journal={Pattern Recognition},
  volume={115},
  pages={107899},
  year={2021},
  publisher={Elsevier}
}

@inproceedings{sun2021explanation,
  title={Explanation-guided training for cross-domain few-shot classification},
  author={Sun, Jiamei and Lapuschkin, Sebastian and Samek, Wojciech and Zhao, Yunqing and Cheung, Ngai-Man and Binder, Alexander},
  booktitle={2020 25th International Conference on Pattern Recognition (ICPR)},
  pages={7609--7616},
  year={2021},
  organization={IEEE}
}

@article{schiller2019relevance,
  title={Relevance-based feature masking: Improving neural network based whale classification through explainable artificial intelligence},
  author={Schiller, Dominik and Huber, Tobias and Lingenfelser, Florian and Dietz, Michael and Seiderer, Andreas and Andr{\'e}, Elisabeth},
  year={2019}
}

@article{zunino2021excitation,
  title={Excitation dropout: Encouraging plasticity in deep neural networks},
  author={Zunino, Andrea and Bargal, Sarah Adel and Morerio, Pietro and Zhang, Jianming and Sclaroff, Stan and Murino, Vittorio},
  journal={International Journal of Computer Vision},
  volume={129},
  number={4},
  pages={1139--1152},
  year={2021},
  publisher={Springer}
}

@article{ismail2021improving,
  title={Improving deep learning interpretability by saliency guided training},
  author={Ismail, Aya Abdelsalam and Corrada Bravo, Hector and Feizi, Soheil},
  journal={Advances in Neural Information Processing Systems},
  volume={34},
  pages={26726--26739},
  year={2021}
}

@inproceedings{ha2019improvement,
  title={Improvement in deep networks for optimization using explainable artificial intelligence},
  author={ha Lee, Jin and hee Shin, Ik and gu Jeong, Sang and Lee, Seung-Ik and Zaheer, Muhamamad Zaigham and Seo, Beom-Su},
  booktitle={2019 International Conference on Information and Communication Technology Convergence (ICTC)},
  pages={525--530},
  year={2019},
  organization={IEEE}
}

@article{bargal2021guided,
  title={Guided zoom: Zooming into network evidence to refine fine-grained model decisions},
  author={Bargal, Sarah Adel and Zunino, Andrea and Petsiuk, Vitali and Zhang, Jianming and Saenko, Kate and Murino, Vittorio and Sclaroff, Stan},
  journal={IEEE Transactions on Pattern Analysis and Machine Intelligence},
  volume={43},
  number={11},
  pages={4196--4202},
  year={2021},
  publisher={IEEE}
}

@inproceedings{zunino2021explainable,
  title={Explainable deep classification models for domain generalization},
  author={Zunino, Andrea and Bargal, Sarah Adel and Volpi, Riccardo and Sameki, Mehrnoosh and Zhang, Jianming and Sclaroff, Stan and Murino, Vittorio and Saenko, Kate},
  booktitle={Proceedings of the IEEE/CVF Conference on Computer Vision and Pattern Recognition},
  pages={3233--3242},
  year={2021}
}

@inproceedings{selvaraju2019taking,
  title={Taking a hint: Leveraging explanations to make vision and language models more grounded},
  author={Selvaraju, Ramprasaath R and Lee, Stefan and Shen, Yilin and Jin, Hongxia and Ghosh, Shalini and Heck, Larry and Batra, Dhruv and Parikh, Devi},
  booktitle={Proceedings of the IEEE/CVF international conference on computer vision},
  pages={2591--2600},
  year={2019}
}

@inproceedings{selvaraju2017grad,
  title={Grad-cam: Visual explanations from deep networks via gradient-based localization},
  author={Selvaraju, Ramprasaath R and Cogswell, Michael and Das, Abhishek and Vedantam, Ramakrishna and Parikh, Devi and Batra, Dhruv},
  booktitle={Proceedings of the IEEE international conference on computer vision},
  pages={618--626},
  year={2017}
}

@article{mahapatra2021interpretability,
  title={Interpretability-driven sample selection using self supervised learning for disease classification and segmentation},
  author={Mahapatra, Dwarikanath and Poellinger, Alexander and Shao, Ling and Reyes, Mauricio},
  journal={IEEE transactions on medical imaging},
  volume={40},
  number={10},
  pages={2548--2562},
  year={2021},
  publisher={IEEE}
}

@inproceedings{sun2022utilizing,
  title={Utilizing Explainable AI for improving the Performance of Neural Networks},
  author={Sun, Huawei and Servadei, Lorenzo and Feng, Hao and Stephan, Michael and Santra, Avik and Wille, Robert},
  booktitle={2022 21st IEEE International Conference on Machine Learning and Applications (ICMLA)},
  pages={1775--1782},
  year={2022},
  organization={IEEE}
}

@inproceedings{hu-etal-2016-harnessing,
    title = "Harnessing Deep Neural Networks with Logic Rules",
    author = "Hu, Zhiting  and
      Ma, Xuezhe  and
      Liu, Zhengzhong  and
      Hovy, Eduard  and
      Xing, Eric",
    editor = "Erk, Katrin  and
      Smith, Noah A.",
    booktitle = "Proceedings of the 54th Annual Meeting of the Association for Computational Linguistics (Volume 1: Long Papers)",
    month = aug,
    year = "2016",
    address = "Berlin, Germany",
    publisher = "Association for Computational Linguistics",
    url = "https://aclanthology.org/P16-1228",
    doi = "10.18653/v1/P16-1228",
    pages = "2410--2420",
}

@article{apicella2018integration,
  title={Integration of context information through probabilistic ontological knowledge into image classification},
  author={Apicella, Andrea and Corazza, Anna and Isgr{\`o}, Francesco and Vettigli, Giuseppe},
  journal={Information},
  volume={9},
  number={10},
  pages={252},
  year={2018},
  publisher={MDPI}
}

@inproceedings{fukui2019attention,
  title={Attention branch network: Learning of attention mechanism for visual explanation},
  author={Fukui, Hiroshi and Hirakawa, Tsubasa and Yamashita, Takayoshi and Fujiyoshi, Hironobu},
  booktitle={Proceedings of the IEEE/CVF conference on computer vision and pattern recognition},
  pages={10705--10714},
  year={2019}
}

@inproceedings{mitsuhara2019embedding,
  author       = {Masahiro Mitsuhara and
                  Hiroshi Fukui and
                  Yusuke Sakashita and
                  Takanori Ogata and
                  Tsubasa Hirakawa and
                  Takayoshi Yamashita and
                  Hironobu Fujiyoshi},
  editor       = {Giovanni Maria Farinella and
                  Petia Radeva and
                  Jos{\'{e}} Braz and
                  Kadi Bouatouch},
  title        = {Embedding Human Knowledge into Deep Neural Network via Attention Map},
  booktitle    = {Proceedings of the 16th International Joint Conference on Computer
                  Vision, Imaging and Computer Graphics Theory and Applications, {VISIGRAPP}
                  2021, Volume 5: VISAPP, Online Streaming, February 8-10, 2021},
  pages        = {626--636},
  publisher    = {{SCITEPRESS}},
  year         = {2021},
  url          = {https://doi.org/10.5220/0010335806260636},
  doi          = {10.5220/0010335806260636}
}

@inproceedings{liu2023icel,
  title={ICEL: Learning with Inconsistent Explanations},
  author={Liu, Biao and Wu, Xiaoyu and Yuan, Bo},
  booktitle={ICASSP 2023-2023 IEEE International Conference on Acoustics, Speech and Signal Processing (ICASSP)},
  pages={1--5},
  year={2023},
  organization={IEEE}
}

@inproceedings{zhou2016learning,
  title={Learning deep features for discriminative localization},
  author={Zhou, Bolei and Khosla, Aditya and Lapedriza, Agata and Oliva, Aude and Torralba, Antonio},
  booktitle={Proceedings of the IEEE conference on computer vision and pattern recognition},
  pages={2921--2929},
  year={2016}
}

@inproceedings{teso2019explanatory,
  title={Explanatory interactive machine learning},
  author={Teso, Stefano and Kersting, Kristian},
  booktitle={Proceedings of the 2019 AAAI/ACM Conference on AI, Ethics, and Society},
  pages={239--245},
  year={2019}
}

@article{anders2022finding,
  title={Finding and removing Clever Hans: using explanation methods to debug and improve deep models},
  author={Anders, Christopher J and Weber, Leander and Neumann, David and Samek, Wojciech and M{\"u}ller, Klaus-Robert and Lapuschkin, Sebastian},
  journal={Information Fusion},
  volume={77},
  pages={261--295},
  year={2022},
  publisher={Elsevier}
}

@inproceedings{liu-avci-2019-incorporating,
    title = "Incorporating Priors with Feature Attribution on Text Classification",
    author = "Liu, Frederick  and
      Avci, Besim",
    editor = "Korhonen, Anna  and
      Traum, David  and
      M{\`a}rquez, Llu{\'\i}s",
    booktitle = "Proceedings of the 57th Annual Meeting of the Association for Computational Linguistics",
    month = jul,
    year = "2019",
    address = "Florence, Italy",
    publisher = "Association for Computational Linguistics",
    url = "https://aclanthology.org/P19-1631",
    doi = "10.18653/v1/P19-1631",
    pages = "6274--6283",
}

@article{Krizhevsky09,
  title={Learning multiple layers of features from tiny images},
  author={Krizhevsky, A. and Hinton, G.},
  journal={Master's thesis, Department of Computer Science, University of Toronto},
  year={2009},
  publisher={Citeseer}
}

@inproceedings{coates2011analysis,
  title={An analysis of single-layer networks in unsupervised feature learning},
  author={Coates, Adam and Ng, Andrew and Lee, Honglak},
  booktitle={Proceedings of the fourteenth international conference on artificial intelligence and statistics},
  pages={215--223},
  year={2011},
  organization={JMLR Workshop and Conference Proceedings}
}

@inproceedings{xie2020self,
  title={Self-training with noisy student improves imagenet classification},
  author={Xie, Qizhe and Luong, Minh-Thang and Hovy, Eduard and Le, Quoc V},
  booktitle={Proceedings of the IEEE/CVF conference on computer vision and pattern recognition},
  pages={10687--10698},
  year={2020}
}

@inproceedings{hagos2022impact,
  title={Impact of Feedback Type on Explanatory Interactive Learning},
  author={Hagos, Misgina Tsighe and Curran, Kathleen M and Mac Namee, Brian},
  booktitle={International Symposium on Methodologies for Intelligent Systems},
  pages={127--137},
  year={2022},
  organization={Springer}
}

@inproceedings{tan2019efficientnet,
  title={Efficientnet: Rethinking model scaling for convolutional neural networks},
  author={Tan, Mingxing and Le, Quoc},
  booktitle={International conference on machine learning},
  pages={6105--6114},
  year={2019},
  organization={PMLR}
}

@article{lecun1998gradient,
  title={Gradient-based learning applied to document recognition},
  author={LeCun, Yann and Bottou, L{\'e}on and Bengio, Yoshua and Haffner, Patrick},
  journal={Proceedings of the IEEE},
  volume={86},
  number={11},
  pages={2278--2324},
  year={1998},
  publisher={Ieee}
}

@misc{howard2017mobilenets,
      title={MobileNets: Efficient Convolutional Neural Networks for Mobile Vision Applications}, 
      author={Andrew G. Howard and Menglong Zhu and Bo Chen and Dmitry Kalenichenko and Weijun Wang and Tobias Weyand and Marco Andreetto and Hartwig Adam},
      year={2017},
      eprint={1704.04861},
      archivePrefix={arXiv},
      primaryClass={cs.CV}
}

@inproceedings{deng2009imagenet,
  title={Imagenet: A large-scale hierarchical image database},
  author={Deng, Jia and Dong, Wei and Socher, Richard and Li, Li-Jia and Li, Kai and Fei-Fei, Li},
  booktitle={2009 IEEE conference on computer vision and pattern recognition},
  pages={248--255},
  year={2009},
  organization={Ieee}
}

@inproceedings{apicella2023shap,
author       = {Andrea Apicella and Salvatore Giugliano and Francesco Isgr{\`{o}} and Roberto Prevete},
  title        = {SHAP-based Explanations to Improve Classification Systems},
  booktitle    = {Proceedings of the 4th Italian Workshop on Explainable Artificial Intelligence co-located with 22nd International Conference of the Italian Association for Artificial Intelligence(AIxIA 2023), Roma, Italy, November 8, 2023},
  series       = {{CEUR} Workshop Proceedings},
  volume       = {3518},
  pages        = {76--86},
  publisher    = {CEUR-WS.org},
  year         = {2023},

}

@article{apicella2023strategies,
	author = {Apicella, Andrea and Di Lorenzo, Luca and Isgrò, Francesco and Pollastro, Andrea and Prevete, Roberto},
	title = {Strategies to Exploit XAI to Improve Classification Systems},
	year = {2023},
	journal = {Communications in Computer and Information Science},
	volume = {1901 CCIS},
	pages = {147 – 159},
    doi={10.1007/978-3-031-44064-9\_9}
}

@article{montavon2019layer,
  title={Layer-wise relevance propagation: an overview},
  author={Montavon, Gr{\'e}goire and Binder, Alexander and Lapuschkin, Sebastian and Samek, Wojciech and M{\"u}ller, Klaus-Robert},
  journal={Explainable AI: interpreting, explaining and visualizing deep learning},
  pages={193--209},
  year={2019},
  publisher={Springer}
}

@incollection{NIPS2017_7062,
title = {A Unified Approach to Interpreting Model Predictions},
author = {Lundberg, Scott M and Lee, Su-In},
booktitle = {Advances in Neural Information Processing Systems 30},
editor = {I. Guyon and U. V. Luxburg and S. Bengio and H. Wallach and R. Fergus and S. Vishwanathan and R. Garnett},
pages = {4765--4774},
year = {2017},
publisher = {Curran Associates, Inc.},
}

@article{erhan2009,
  title={Visualizing higher-layer features of a deep network},
  author={Erhan, D. and Bengio, Y. and Courville, . and Vincent, P.},
  journal={University of Montreal},
  volume={1341},
  number={3},
  pages={1},
  year={2009}
}

@article{bach2015,
  title={On pixel-wise explanations for non-linear classifier decisions by layer-wise relevance propagation},
  author={Bach, Sebastian and Binder, Alexander and Montavon, Gr{\'e}goire and Klauschen, Frederick and M{\"u}ller, Klaus-Robert and Samek, Wojciech},
  journal={PloS one},
  volume={10},
  number={7},
  pages={e0130140},
  year={2015},
  publisher={Public Library of Science}
}

@inproceedings{binder2016,
  title={Layer-wise relevance propagation for neural networks with local renormalization layers},
  author={Binder, Alexander and Montavon, Gr{\'e}goire and Lapuschkin, Sebastian and M{\"u}ller, Klaus-Robert and Samek, Wojciech},
  booktitle={International Conference on Artificial Neural Networks},
  pages={63--71},
  year={2016},
  address = {Barcelona, Spain},
  organization={Springer}
}

@article{montavon2017_2,
  title={Explaining nonlinear classification decisions with deep taylor decomposition},
  author={Montavon, Gr{\'e}goire and Lapuschkin, Sebastian and Binder, Alexander and Samek, Wojciech and M{\"u}ller, Klaus-Robert},
  journal={Pattern Recognition},
  volume={65},
  pages={211--222},
  year={2017},
  publisher={Elsevier}
}

@inproceedings{zeiler2011,
  title={Adaptive deconvolutional networks for mid and high level feature learning},
  author={Zeiler, M. D. and Taylor, G. W. and Fergus, R.},
  booktitle={Computer Vision (ICCV), 2011 IEEE International Conference on},
  pages={2018--2025},
  year={2011},
  organization={IEEE},
  address = {Barcelona, Spain}
}

@inproceedings{zeiler2014,
  title={Visualizing and understanding convolutional networks},
  author={Zeiler, M. D. and Fergus, R.},
  booktitle={European Conference on Computer Cision},
  pages={818--833},
  year={2014},
  organization={Springer},
  address = {Zurich, Switzerland}
}

@inproceedings{dosovitskiy2016,
  title={Inverting visual representations with convolutional networks},
  author={Dosovitskiy, A. and Brox, T.},
  booktitle={Proceedings of the IEEE Conference on Computer Vision and Pattern Recognition},
  pages={4829--4837},
  year={2016},
  address = {Las Vegas, USA}
}

@article{kakogeorgiou2021evaluating,
  title={Evaluating explainable artificial intelligence methods for multi-label deep learning classification tasks in remote sensing},
  author={Kakogeorgiou, Ioannis and Karantzalos, Konstantinos},
  journal={International Journal of Applied Earth Observation and Geoinformation},
  volume={103},
  pages={102520},
  year={2021},
  publisher={Elsevier}
}

@article{tjoa2022quantifying,
  title={Quantifying explainability of saliency methods in deep neural networks with a synthetic dataset},
  author={Tjoa, Erico and Guan, Cuntai},
  journal={IEEE Transactions on Artificial Intelligence},
  volume={4},
  number={4},
  pages={858--870},
  year={2022},
  publisher={IEEE}
}

@article{islam2022systematic,
  title={A systematic review of explainable artificial intelligence in terms of different application domains and tasks},
  author={Islam, Mir Riyanul and Ahmed, Mobyen Uddin and Barua, Shaibal and Begum, Shahina},
  journal={Applied Sciences},
  volume={12},
  number={3},
  pages={1353},
  year={2022},
  publisher={MDPI}
}

@article{vaswani2017attention,
  title={Attention is all you need},
  author={Vaswani, A},
  journal={Advances in Neural Information Processing Systems},
  year={2017}
}





% \end{thebibliography}
\end{document}